\documentclass[10pt,twocolumn,letterpaper]{article}

\usepackage{cvpr}              

\usepackage[dvipsnames]{xcolor}

\usepackage{multirow}
\usepackage{color}
\usepackage{xcolor}
\usepackage{bigstrut}
\usepackage{bbm}
\usepackage[ruled,vlined]{algorithm2e}
\usepackage{colortbl}
\definecolor{cvprblue}{rgb}{0.21,0.49,0.74}
\usepackage[pagebackref,breaklinks,colorlinks,citecolor=cvprblue]{hyperref}
\usepackage{algpseudocode}
\def\paperID{10778} 
\def\confName{CVPR}
\def\confYear{2025}
\usepackage[accsupp]{axessibility}

\title{From Laboratory to Real World: A New Benchmark Towards Privacy-Preserved Visible-Infrared Person Re-Identification}
\author{Yan Jiang\textsuperscript{1,2}, Hao Yu\textsuperscript{2}, Xu Cheng\textsuperscript{1}\thanks{Corresponding Author}, Haoyu Chen\textsuperscript{2}, Zhaodong Sun\textsuperscript{1,2}, Guoying Zhao\textsuperscript{2}\\
\textsuperscript{1} School of Computer Science, Nanjing University of Information Science and Technology\\ \quad 
\textsuperscript{2} Center for Machine Vision and Signal Analysis, University of Oulu\\
{\tt\small \{jiangyan,xcheng,zhaodong.sun\}@nuist.edu.cn}
\quad{\tt\small \{hao.2.yu,chen.haoyu,guoying.zhao\}@oulu.fi} \\
\tt\small \url{https://github.com/Joey623/L2RW}
}

\begin{document}
\maketitle
\begin{abstract}
Aiming to match pedestrian images captured under varying lighting conditions, visible-infrared person re-identification (VI-ReID) has drawn intensive research attention and achieved promising results. However, in real-world surveillance contexts, data is distributed across multiple devices/entities, raising privacy and ownership concerns that make existing centralized training impractical for VI-ReID. To tackle these challenges, we propose L2RW, a benchmark that brings VI-ReID closer to real-world applications. The rationale of L2RW is that integrating decentralized training into VI-ReID can address privacy concerns in scenarios with limited data-sharing regulation. Specifically, we design protocols and corresponding algorithms for different privacy sensitivity levels. In our new benchmark, we ensure the model training is done in the conditions that: 1) data from each camera remains completely isolated, or 2) different data entities (e.g., data controllers of a certain region) can selectively share the data. In this way, we simulate scenarios with strict privacy constraints which is closer to real-world conditions. Intensive experiments with various server-side federated algorithms are conducted, showing the feasibility of decentralized VI-ReID training. Notably, when evaluated in unseen domains (i.e., new data entities), our L2RW, trained with isolated data (privacy-preserved), achieves performance comparable to SOTAs trained with shared data (privacy-unrestricted). We hope this work offers a novel research entry for deploying VI-ReID that fits real-world scenarios and can benefit the community.
\end{abstract}    
\section{Introduction}
\label{sec:intro}
The growing need for dependable pedestrian identification across all hours has made visible-infrared person re-identification (VI-ReID) a key technology \cite{SYSU,DEEN,CAJ+}. It solves the challenge of identifying individuals in different spectral modalities, such as visible during the day and infrared at night. However, although existing methods \cite{TOPlight,MUN,FDNM,IDKL,PGM,GUR,SDCL,CAL} achieve encouraging advancements, their deployment in real-world scenarios remains limited.

\begin{figure}[t]
    \centering
    \includegraphics[width=0.48\textwidth]{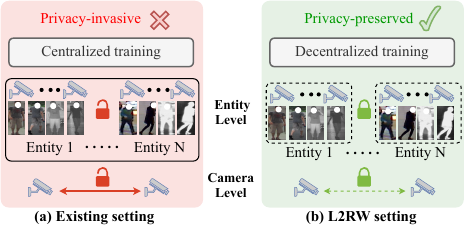}
    \vspace{-2.0em}
    \caption{Illustration of existing setting and the proposed L2RW setting. (a) The existing setting relies on centralized training, allowing unrestricted data sharing between cameras or entities, which brings serious privacy concerns. (b) Unlike this, our L2RW framework introduces camera-level and entity-level privacy regulation, simulating real-world privacy-preserved VI-ReID scenes.
    }
    \label{fig:1}
    \vspace{-1.0em}
\end{figure}

In our view, the current experimental setting of the VI-ReID in the field is idealized and impedes its advancement to real-world applications. Specifically, current approaches \cite{AIO,Unireid,zhu2024cross,pan2024unified,qian2025visible,MMM,GUR,PGM,jiang2025dsaf} rely predominantly on centralized training, where data from multiple sources is aggregated on a central platform. While beneficial for learning modality-shared representations, they neglect practical constraints, especially in privacy and ownership concerns. In real-world scenarios, data is stored in a distributed manner, surveillance images are subject to strict privacy restrictions, etc. This centralized training not only complicates data sharing but also increases the risk of data leakage, bringing serious privacy concerns in real-world deployment.

Based on the above observation, we argue that it is essential to establish a benchmark that protects data privacy while still enabling effective VI-ReID. Therefore, we propose L2RW (\textbf{L}aboratory \textbf{to} \textbf{R}eal \textbf{W}orld), a benchmark designed to simulate real-world VI-ReID scenarios with privacy preservation at its core. In L2RW, privacy is not an afterthought but a fundamental consideration. However, in reality, determining which data should be protected for privacy remains a challenging issue. For example, \textbf{1)} pedestrian images captured by surveillance cameras are typically independent and without direct data sharing across cameras. \textbf{2)} The entity, e.g., the data owner of a certain region, has the right to access all data. When different entities collaborate, should their data be isolated or kept shared? 

This complexity brings us to a natural question in designing the L2RW benchmark for real-world VI-ReID: \textit{How can we define a proper and feasible privacy framework to adapt to different privacy levels?} Thus, we design three protocols, i.e., Camera Independence (CI), Entity Independence (EI), and Entity Sharing (ES), for flexible privacy level control. For CI, all camera data remains isolated, simulating the highest privacy level where no data sharing occurs across every single camera as shown in the camera level of Fig.~\ref{fig:1}(b). Regarding a more relaxed privacy level in which all entity data is isolated while data can be shared inside each data entity, we propose the EI protocol as shown in the entity level of Fig.~\ref{fig:1}(b). When all data entities can be shared for centralized training, we name this protocol ES as shown in Fig.~\ref{fig:1}(a), which is the commonly used protocol in previous Re-ID methods. In the ES protocol, existing VI-ReID methods \cite{AGW,CAJ,Lba,DEEN,DNS} can be seamlessly applied. However, note that these algorithms are designed for a closed-world setting, where training and testing are conducted on the same entity. To avoid this, in our ES protocol, we train on multiple shared entities but directly test on an unseen entity.

Evidently, existing VI-ReID methods cannot be directly applied under these privacy-preserving protocols (CI and EI) due to many technical obstacles: 1) In CI, a single camera may consist of only visible, only infrared, or both modalities, this modality mismatch between different cameras makes the training extremely challenging, leading to a \textbf{modality incomplete} issue. 2) In EI, since pedestrians vary across different entities, this non-sharing approach introduces the \textbf{identity missing} issue. 3) In both the CI and EI protocols, differences in data distribution across cameras or entities inevitably bring the \textbf{domain shift} issue.

To address the above challenges and make the L2RW benchmark actionable, we provide a novel algorithm customized for this task, named Decentralized Privacy-Preserved Training (DPPT) as a baseline solution. Inspired by the substantial success of federated learning \cite{FedAVG,FedProx,Fednova,MOON,FCCL,MarsFL,FCCL+}, our DPPT employs decentralized training to ensure privacy protection. To solve the \textit{modality incomplete} issue, DPPT converts the existing two-stream architecture into a one-stream architecture and modifies the current sampling way to eliminate the need for modality distinction. To handle the \textit{identity missing} and \textit{domain shift} issues, we design a memory rectification bank (MRB) that selects the top K features closest to the center point within each class as representatives. By averaging these top K features, we correct the center point to better reflect the current dominant class distribution. Then, MRB averages the dominant centers across clients and stores them in a global memory, which serves as a fair convergence point. This enables different clients to optimize towards the same goal. Generally, the contributions of this paper are summarized as follows: 

\begin{itemize}
    \item We propose a benchmark named L2RW to simulate privacy-preserving real-world VI-ReID scenarios. Within the L2RW benchmark, we design three protocols, i.e., CI (Camera Independence), EI (Entity Independence) and ES (Entity Sharing). This simulates scenarios with different privacy constraints, bringing VI-ReID evaluation closer to real-world conditions.
    \item We analyze the challenges encountered under the CI and EI protocols and propose a unified method named DPPT, which is the first work that handles privacy concerns for VI-ReID in a decentralized manner.
    \item In our L2RW benchmark, unlike existing methods that validate solely on a single dataset, we merge several existing datasets and conduct the evaluation in a cross-domain manner to simulate the real-world scenarios.
    \item Extensive experiments on three public VI-ReID datasets confirm the feasibility of decentralized training in L2RW, with our method achieving significant improvement on various federated learning baselines under CI. We also show that DPPT achieves performance under EI comparable to that of ES.
\end{itemize}

\section{Related Work}
\label{sec:related}

\subsection{Visible-Infrared Person ReID}
VI-ReID aims to match visible and infrared images of the same pedestrian captured by non-overlapped cameras. Many studies have emerged recently, achieving remarkable progress. For instance, Zhang \textit{et al.} \cite{DEEN} explored diverse modality-shared features to mine significant cross-modality patterns. Jiang \textit{et al.} \cite{DNS} designed domain shifting (DNS) that augments modality-specific and modality-shared representation, thereby regulating the model to concentrate on consistency between modalities. Ren \textit{et al.} proposed a novel implicit discriminative knowledge learning (IKDL) network to discover identity-aware salient information for aligning visible and infrared modalities.
Some auxiliary-based methods \cite{SGIEL,MUN,MMN} have also been developed to achieve modality alignment with the help of generated auxiliary modalities. Feng \textit{et al.} \cite{SGIEL} proposed a shape-guided diverse feature learning framework (SGIEL) that employs the body shape as the auxiliary information for modality alignment. Yu \textit{et al.} \cite{MUN} proposed a modality unifying network (MUN) that constructed a robust auxiliary modality by intra- and inter-modality learners.

However, all the above methods are achieved under idealized lab settings, which neglect privacy constraints and heterogeneous environments pose significant challenges in reality. More importantly, these methods require paired visible and infrared images, making privacy-preserved implementation infeasible, especially under CI, where modality information is unknown. In addition, existing works neglect the research on unseen environments, struggling to generalize effectively in real-world scenes. Therefore, we propose L2RW, offering a novel practical and privacy-preserving benchmark for bringing VI-ReID closer to real world.


\subsection{Federated Learning}

Federated Learning is a technology that enables multiple devices to collaborative learning without sharing private data. The pioneering work, FedAvg \cite{FedAVG}, averages the gradients of locally trained models and redistributes them to local clients for further training. Based on FedAvg, Li \textit{et al.} \cite{FedProx} proposed FedProx, introducing a proximal term to ensure the convergence of the network in federated environments. Subsequently, Li \textit{et al.} \cite{MOON} designed MOON, which utilizes the modality representation similarity to correct the local training of clients.
These methods brought new insights to the community, inspiring numerous impressive works that solve data heterogeneity \cite{FL, vitfl, FedSAM,FedProto} and model heterogeneity \cite{FCCL, FedHeNN, HFL} in federated learning. This progress provides solid technical support for our L2RW.

However, despite these advancements, existing federated learning methods cannot be directly applied to VI-ReID as they cannot handle specific issues such as modality incomplete, identity missing, and domain shift encountered in real-world scenarios. 

\subsection{Domain Generalization}

Domain generalization (DG) uses multiple seen domains to train a model that generalizes well to unseen domains. This technology aims to address the key issue of current deep learning methods, which heavily rely on the assumption that training and testing sets are independently and identically distributed. In practice, new information often arrives in continuous portions, leading to a domain gap between newly acquired data and the original data. This gap makes it challenging for the trained model to perform well on unseen data. Numerous works have emerged to extract domain-invariant features via contrastive loss \cite{DG-contrast1,DG-contrast2,ballas2024multi}, adversarial learning \cite{DG-Adver,DG-Adver1,MADG}, causal learning \cite{DG-casual2, DG-casual1,sun2023unbiased}, etc. Despite their success, these methods require centralized data from source domains and assume that data within the same domain share the same distribution. These not only limit the applicability of these methods in decentralized settings but also increase the risk of privacy leakage.

Conversely, our designed EI protocol in the L2RW benchmark eliminates the need to share source domain data, addressing privacy concerns. This also shows the feasibility of privacy domain generalization in VI-ReID and shows its potential for secure and effective deployment in reality.

\section{Methodology}
In this section, we describe the proposed L2RW benchmark in detail. In brief, Sec.~\ref{sec:3.1} will present the overview of the L2RW. Then, we elaborate on the three protocols we designed, which are grounded in real-world scenarios: \textit{camera independence} (CI) in Sec.~\ref{sec:3.2}, \textit{entity independence} (EC) and \textit{entity sharing} (ES) in Sec.~\ref{sec:3.3}. Finally, we introduce our proposed decentralized privacy-preserved training (DPPT) in Sec.~\ref{sec:3.4}.

\begin{figure*}[t]
    \centering
    \includegraphics[width=1.0\textwidth]{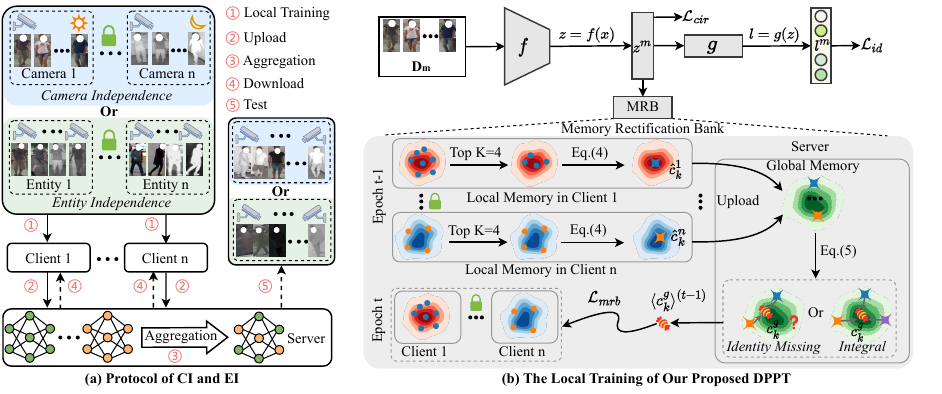}
    \vspace{-2.0em}
    \caption{Protocols and proposed method. (a) are the protocols of camera independence (CI) and entity independence (EI) for privacy-preserved VI-ReID. Another our proposed protocol entity sharing (ES) is omitted in the figure, cause its training way is the same as that used in existing VI-ReID methods. (b) is the local training of our proposed decentralized privacy-preserved training (DPPT).}
    \label{fig:2}
\end{figure*}

\subsection{Overview}
\label{sec:3.1}
The overall pipeline of the L2RW benchmark is shown in Fig.~\ref{fig:2}(a). Let $D_m=\{(x_i, y_i)\}_{i=1}^{N_m}$ denote the $\textit{m-th}$ local private data, where $x_i$ and $y_i$ denote the pedestrian images and corresponding label, and $N_m$ is the local data scale. The overall training set is denoted as $D=\{D_1, D_2, ..., D_n\}$. In the realistic VI-ReID, each private dataset $D_m$ can represent the data collected by a single camera or the surveillance data from within a specific entity. The overall process of the L2RW can be roughly summarized into five steps. \textcircled{1} \textbf{Local Training:} Each private data is trained by a local model $\theta^l$ (client). \textcircled{2} \textbf{Upload:} Each client model weight is uploaded to the server. \textcircled{3} \textbf{Aggregation:} The uploaded local model weights are aggregated to obtain the global model weight $\theta^g$. \textcircled{4} \textbf{Download:} The global model weight $\theta^g$ is downloaded by the clients for next local training. By repeating step \textcircled{1} to \textcircled{4} until convergence, execute \textcircled{5} \textbf{Test:} test seen domains or unseen domains according to the protocol.

\subsection{Benchmark Protocols}
Considering real-world conditions, we design three protocols, i.e., \textit{camera independence} (CI), \textit{entity independence} (EI) and \textit{entity sharing} (ES), to simulate different privacy-restricted scenarios.

\subsubsection{Protocol 1: Camera Independence (CI)}
\label{sec:3.2}
\textbf{CI:} \textit{all camera data remains isolated, which simulates the most strict setting where no data sharing occurs across every single device.} 

It solves the concerns about data privacy, ownership regulations, or security policies in reality. However, existing VI-ReID methods cannot be directly applied under CI due to the \textit{modality incomplete} issue. Specifically, current VI-ReID methods require paired visible and infrared images of the same identity as input during training, which is achieved through the \textit{PK} strategy. $P$ identities with $K$ visible and $K$ infrared images are randomly sampled from the training set (i.e., mini-batch=$2\times P \times K$). These methods typically use a dual-stream architecture to extract modality-specific features in the shallow layers and modality-shared features in the deeper layers, relying on modality information as a prerequisite. Consequently, these requirements make existing VI-ReID methods incompatible with the CI protocol due to the challenges posed by \textit{modality incomplete}.

To solve these issues, we convert the two-stream structure into one-stream architecture and only sample images according to identity rather than identity and modality. Specifically, we randomly sample $K$ images from $P$ identities in the \textit{m-th} local private data $D_m$, where the $K$ images can be visible, infrared, or a mixed of both. This eliminates the reliance on the paired visible and infrared images of the same subject as the input, thereby addressing the \textit{modality incomplete}. In addition, we also employ channel augmentation (CA) \cite{CAJ} to destroy the color information by randomly selecting one channel (R, G, or B) to replace the other channels, which reduces the cross-modality gap across clients.

\subsubsection{Protocol 2\&3: Entity Independence and Entity Sharing (EI\&ES)}
\label{sec:3.3}
In real-world scenarios, it is often feasible for an entity (data controller) to access all data within its designated region or with other data entities, which is a more relaxed privacy regulation compared to CI. Therefore, based on varying levels of mutual trust and privacy agreements, we categorize entity-level privacy regulation into two protocols: \textbf{Entity Independence (EI)} and \textbf{Entity Sharing (ES)}.

\textbf{EI:} \textit{Data from different entities is kept isolated during training, and the model is trained using a decentralized way. Then evaluate the trained model on an unseen entity to assess its generalization ability.}

EI ensures that data sharing across entities is entirely avoided, aligning with real-world privacy requirements and ensuring data security. It also guarantees that knowledge is still transferred across entities without exposing raw data, providing a practical and secure solution for VI-ReID in real-world deployments.

\textbf{ES:} \textit{Data from different entities is shared during training, and the model is evaluated on an unseen entity to assess its generalization ability.} 

In such an ES protocol, existing VI-ReID methods can be seamlessly applied. However, note that these methods' evaluating protocols are often designed on the same dataset, where the data distribution is assumed to be consistent. Meanwhile, in our provided ES protocol, we merge several existing datasets and conduct the evaluation in a cross-domain manner to simulate real-world scenarios. To the best of our knowledge, this has not been explored and remains an open issue. 

We reproduce several VI-ReID methods \cite{Lba,AGW,CAJ,DEEN,DNS} under this protocol, which will be discussed in Sec.~\ref{sec:4.3}. Additionally, we provide a baseline using Empirical Risk Minimization (ERM) \cite{ERM}. It merges data from all training domains to learn domain-invariant knowledge and can be formulated as follows:
\begin{equation}
\begin{aligned}
    \mathcal{L}_{ERM} = \min\limits_{f,g}\frac{1}{|\mathcal{V}|}\sum\limits_{V_i\in \mathcal{V}}\mathcal{R}^{V_i}(f, g),
    \label{eq:1}
\end{aligned}
\end{equation}
where $f$ and $g$ are the encoder and classifier, respectively. $V_i$ is the \textit{i-th} entity, and $\mathcal{V}$ is a set containing all entities. The empirical risk function $\mathcal{R}^{V_i}(f, g)$ for a given entity $V_i$ is defined by:

\begin{equation}
\begin{aligned}
    \mathcal{R}^{V_i}(f, g) \triangleq \mathbb{E}_{(x_j, y_j) \sim V_i} \mathcal{L} ((x_j;f,g),y_j),
    \label{eq:2}
\end{aligned}
\end{equation}
where our baseline loss function $\mathcal{L}(\cdot,\cdot)$ includes the identity loss and circle loss. It is worth noting that this protocol is fully compatible with existing methods designed for laboratory settings, as it does not change their training paradigm. The loss function $\mathcal{L}(\cdot,\cdot)$ can be replaced with the specific loss functions employed by those methods, ensuring seamless integration with their frameworks. Our protocol provides a fair and comprehensive evaluation of the generalization capability of existing VI-ReID approaches. 


\subsection{DPPT Method}
\label{sec:3.4}
To address the challenges in CI and EI, we propose decentralized privacy-preserved training (DPPT). DPPT consists of five steps aimed at addressing the challenges encountered in these protocols that cannot be effectively resolved through existing VI-ReID methods:

We first outline the challenges encountered under CI and EI, which cannot be handled by existing VI-ReID methods:
\begin{itemize}
    \item \textbf{Domain shift:} $P_i(x|y)\neq P_j(x|y)$. Pedestrian images on different clients will exhibit the non-iid (Independent and identically distributed) issue, where the corresponding feature distributions are different across different clients. 
    \item \textbf{Modality incomplete:} For an individual client, pedestrian images can fall into three scenarios: only visible, only infrared, or both visible and infrared. 
    \item \textbf{Identity missing:} $P_i(y)\neq P_j(y)$. The subject is hard to appear in all cameras or entities.
\end{itemize}

Therefore, we propose the DPPT, which consists of five steps that we will introduce one by one.

\textcircled{1} \textbf{Local Training:} The process of client local training is shown in Fig.~\ref{fig:2}(b). The private data are fed into the encoder and get the $l_2$-normalized feature embedding $z$ and logits $l$. The identity loss $\mathcal{L}_{id}$ and circle loss $\mathcal{L}_{cir}$ are employed as the baseline loss functions (provided in the supplementary material) to ensure the learned features and logits are identity-related.

However, this training objective design cannot address the domain shift and identity missing, as the model tends to overfit the current private data in the client training and forget previously learned knowledge after download. Driven by this, we propose a memory rectification bank (MRB) that enables different clients to optimize towards the same goal.

\noindent\textbf{MRB.} Specifically, we design a local memory $\mathcal{M}^{l}$ to store $l_2$-normalized feature embedding $z$ in each client. We first calculate the mean vector $c_k$ that store features belonging to the same class via Eq.~\ref{eq:3}:

\begin{equation}
\begin{aligned}
    c_k^m = \frac{1}{|S_k^m|} \sum_{(z_i^m, y_i^m)\in S_k^m} z_i^m,
    \label{eq:3}
\end{aligned}
\end{equation}
where $S_k^m$ denotes the set of samples annotated with class $k$ in the \textit{m-th} client. The local memory can be denoted as $\mathcal{M}^{l}_{m}=\{c_1^m, c_2^m, ..., c_{I_m}^{m}\}$. Each center can be viewed as a representative of each identity, containing client-specific information. By aggregating the center from different clients, a representative encompassing all clients can be obtained without leaking the original data. However, the modality and number of images for each pedestrian within a camera are unknown, and the centers for different individuals are generated from varying sample numbers, which undermines fairness between identities. Additionally, images affected by occlusions or environmental interference contain lower semantic information, which can distort the center calculation. Therefore, our MRB adjusts the centers by averaging the top $K$ closest feature embeddings to the center, effectively filtering out low-information features and providing a fairer representation for each identity. It is defined as:
\begin{equation}
\hat{c}_k^m =
\begin{cases}
    c_k^m, & \text{if } |S_k^m| < K, \\[2ex]
    \displaystyle \frac{1}{K} \sum \text{min}_{K} \, d(z_i^m, c_k^m), & \text{if } |S_k^m| \geq K,
\end{cases}
\label{eq:4}
\end{equation}
where $d(\cdot, \cdot)$ is the Euclidean distance and the local memory is updated as $\mathcal{M}_m^l=\{\hat{c}_1^m, \hat{c}_2^m, ..., \hat{c}_{I_m}^m\}$, $I_m$ is the identity number in client $m$. 
The local memory is uploaded to the server and aggregated through a global memory $\mathcal{M}^g$, as defined below:
\begin{equation}
\begin{aligned}
     \mathcal{M}^g = \left\{c_i^g = \frac{\sum_{m=1}^{n}\mathbbm{1}_{\{\hat{c}_i^{m}\neq0\}}\cdot\hat{c}_i^{m}}{\sum_{m=1}^{n}\mathbbm{1}_{\{\hat{c}_i^{m}\neq0\}}} \middle|i=1,2,\dots, I\right\},
    \label{eq:5}
\end{aligned}
\end{equation}
where $n$ is the number of clients. $I$ is the total identity number, and $I_m\leq I$, thereby handling the \textit{identity missing} issue. $\mathbbm{1}$ denotes the indicator function. The global memory contains information from different domains while ensuring no data leakage. In the next round of local training, we use the memory rectification bank loss $\mathcal{L}_{mrb}$ to guide the optimization of the locally updated memory towards the global memory, thereby relieving domain shift. The $\mathcal{L}_{mrb}$ is defined as follows:
\begin{equation}
\begin{aligned}
    \mathcal{L}_{mrb} = \frac{1}{N_m}\sum_{i=1}^{N_m}(1-\frac{\langle z_i^{m}\rangle^{(t)} \cdot \langle c_{y_i}^{g}\rangle^{(t-1)}}{||\langle z_i^{m}\rangle^{(t)}||\times ||\langle c_{y_i}^{g}\rangle^{(t-1)}||}),
    \label{eq:6}
\end{aligned}
\end{equation}
where $\langle z_i^{m}\rangle^{(t)}$ is the \textit{i-th} $l2$-normalized feature embedding in the \textit{t-th} epoch on the \textit{m-th} client. $\langle c_{y_i}^{g}\rangle^{(t-1)}$ is the global center of identity $y_i$ in the last epoch $t-1$, which contains the knowledge from the client $m$ and the other clients without leaking raw data. $\mathcal{L}_{mrb}$ pulls together the local memory and the global memory, addressing the domain shift and identity missing. The total training loss of the $m-th$ client can be written as:

\begin{equation}
\begin{aligned}
    \mathcal{L}_{all} = \mathcal{L}_{id} + \mathcal{L}_{cir} + \lambda\mathcal{L}_{mrb},
    \label{eq:7}
\end{aligned}
\end{equation}
where $\lambda$ is the loss balance factor.

\textcircled{2} \textbf{Upload:} The client model parameters $\theta^l$ are upload to the server, denoted by $\{\theta_1^l, ..., \theta_n^l\}$.

\textcircled{3} \textbf{Aggregation:} The uploaded local model parameters are aggregated to obtain the global model parameters $\theta^g$:
\begin{equation}
\begin{aligned}
    \theta^g = \sum_{i=1}^{n}\frac{N_i}{N_{total}}\theta^l_i,
    \label{eq:8}
\end{aligned}
\end{equation}
where $N_{total}$=$\sum_{i=1}^{n}N_i$ is the total number of samples across all clients.

\textcircled{4} \textbf{Download:} The global model parameters $\theta^g$ is downloaded by the clients for next local training.

\textcircled{5} \textbf{Test:} After training, test seen domains or unseen domains according to the CI or EI protocols. The overall process is shown in Algorithm \ref{alg1}.

\SetKwComment{Comment}{/* }{*/}
\RestyleAlgo{ruled}
\begin{algorithm}[t]
\caption{DPPT}\label{alg1}
\SetKwInOut{KwData}{Input}
\SetKwInOut{KwResult}{Output}
\KwData{The number of clients $n$, initial global model parameters $\theta^g$, private local model parameters $\{\theta^l_1, \theta^l_2,...,\theta^l_n\}$, local private datasets $\{D_1, D_2,...,D_n\}$, and the number of training epochs $E$.}
\KwResult{The global model parameters $\theta^g$ for test.}
\For {$e = 1$ to $E$}{
    \For{$m=1, 2, \dots, n$}{
        $\theta_m^{l}, \mathcal{M}_m^{l} \gets \textbf{ClientTraining($\theta^g, D_m, \mathcal{M}^{g}$)}$\;
    }
    \tcp{Upload and aggregate the local models}
    $\theta^{g} \gets \sum_{i=1}^{n}\frac{|D_i|}{\sum_{j=1}^{n}|D_j|}\theta_i^l$ \;
    \tcp{Upload and aggregate the local memory banks}
    $\mathcal{M}^g \gets \{c_i^g = \frac{\sum_{m=1}^{n}\mathbbm{1}_{\{\hat{c}_i^{m}\neq0\}}\cdot\hat{c}_i^{m}}{\sum_{m=1}^{n}\mathbbm{1}_{\{\hat{c}_i^{m}\neq0\}}}|i=1,2,..., I\}$\;
}
\textbf{ClientTraining($\theta^g, D_m,\mathcal{M}^g$):}\\
$\theta_m^l \gets \theta^g $ \tcp{Download the global model to the local model}
\For{$(x_i, y_i)\in D_m$}{
    $z_i=f_m(x_i), l_i=g_m(z_i)$\;
    $\mathcal{L}_{cir} \gets (z_i, y_i)$, $\mathcal{L}_{id} \gets (l_i, y_i)$ \;
    $\mathcal{M}_m^l \gets z_i$ via Eq.(\ref{eq:3},\ref{eq:4})\;
    $\mathcal{L}_{mrb}\gets (\mathcal{M}_m^l, \mathcal{M}^g)$ via Eq.(\ref{eq:6})\;
    $\mathcal{L}=\mathcal{L}_{id}+\mathcal{L}_{cir}+\lambda\mathcal{L}_{mrb}$\;
    $\theta^l_m\gets \theta^l_m - \eta\nabla\mathcal{L}$\;
}
return $\theta_m^l, \mathcal{M}_m^l$\;

\end{algorithm}
\section{Experiment}
\label{experiment}

\begin{table*}[t]
  \centering
  \caption{Evaluation of our DPPT under CI protocol. \textit{Note that existing VI-ReID methods cannot directly be applied to the CI protocol as their frameworks are designed for data-shared learning.} Thus, we implemented four classic federated learning algorithms as baselines to verify the efficacy of our method: FedProx \cite{FedProx}, Fednova \cite{Fednova}, Moon \cite{MOON}, and FedAvg \cite{FedAVG}. Evaluating metrics rank-1(\%), rank-10(\%), mAP(\%), and mINP(\%) are reported. AGW$^\dagger$ and DNS$^\dagger$ are the reproduced VI-ReID method that removes the modality information.}
  \vspace{-1.0em}
  \resizebox{\textwidth}{!}{
  \setlength\tabcolsep{11.0pt}
    \begin{tabular}{c|cccc|cccc|cccc}
    \hline
    \multirow{2}[4]{*}{Methods} & \multicolumn{4}{c|}{SYSU-MM01 \cite{SYSU}} & \multicolumn{4}{c|}{RegDB \cite{RegDB}}    & \multicolumn{4}{c}{LLCM \cite{DEEN}} \bigstrut\\
\cline{2-13}          & r=1 $\uparrow$   & r=10 $\uparrow$  & mAP $\uparrow$   & mINP $\uparrow$  & r=1 $\uparrow$   & r=10 $\uparrow$  & mAP $\uparrow$   & mINP $\uparrow$  & r=1 $\uparrow$   & r=10 $\uparrow$  & mAP $\uparrow$   & mINP $\uparrow$ \bigstrut[t]\\
    \hline
    FedProx \cite{FedProx} & 25.90  & 68.83  & 27.48  & 17.18  & 25.62  & 44.19  & 25.96  & 17.80  & 23.72  & 53.95  & 30.59  & 27.69  \bigstrut[t]\\
    +AGW$^\dagger$ \cite{AGW}  & 21.50  & 62.59  & 23.07  & 13.89  & 20.82  & 38.94  & 21.57  & 14.35  & 24.65  & 55.98  & 31.62  & 28.45 \\  
    +DNS$^\dagger$ \cite{DNS}  & 36.11  & 78.29  & 35.22  & 22.02  & 46.75  & 69.27  & 43.06  & 28.99  & 26.35  & 57.00  & 32.88  & 29.54  \\
    \rowcolor[rgb]{0.9, 0.9, 0.98}
    +DPPT (Ours) & \textbf{38.16}  & \textbf{81.54}  & \textbf{38.15}  & \textbf{25.17}  & \textbf{51.33}  & \textbf{71.43}  & \textbf{48.93}  & \textbf{36.15}  & \textbf{27.46}  & \textbf{59.33}  & \textbf{34.60}  & \textbf{31.44}  \\
    \hline
    Fednova \cite{Fednova} & 29.15  & 74.00  & 31.34  & 20.77  & 20.50  & 35.00  & 22.15  & 16.41  & -     & -     & -     & - \bigstrut[t]\\
    +AGW$^\dagger$ \cite{AGW}  & 22.00  & 64.15  & 23.52  & 13.54  & 13.83  & 25.47  & 15.92  & 12.21  & -  & -  & -  & - \\  
    +DNS$^\dagger$ \cite{DNS}  & 40.79  & 83.14  & 41.01  & 27.87  & 46.70  & 68.15  & 43.40  & 29.92  & -     & -     & -     & - \\
    \rowcolor[rgb]{0.9, 0.9, 0.98}
    +DPPT (Ours) & \textbf{50.37}  & \textbf{88.92}  & \textbf{48.67}  & \textbf{33.65}  & \textbf{59.67}  & \textbf{78.36}  & \textbf{55.17}  & \textbf{41.12}  & -     & -     & -     & - \\
    \hline
    Moon \cite{MOON}  & 26.88  & 71.65  & 29.54  & 19.55  & 19.96  & 34.19  & 21.54  & 15.93  & 25.02  & 57.58  & 32.34  & 29.25  \bigstrut[t]\\
    +AGW$^\dagger$ \cite{AGW}  & 20.79  & 62.57  & 22.44  & 12.80  & 11.19  & 20.56  & 13.44  & 10.46  & 23.60  & 55.98  & 31.62  & 28.45 \\  
    +DNS$^\dagger$ \cite{DNS}  & 38.18  & 81.31  & 38.94  & 26.53  & 43.60  & 65.46  & 40.33  & 27.87  & 29.55  & 60.82  & 36.64  & 33.52  \\
    \rowcolor[rgb]{0.9, 0.9, 0.98}
    +DPPT (Ours) & \textbf{46.78}  & \textbf{87.40}  & \textbf{45.99}  & \textbf{31.74}  & \textbf{53.22}  & \textbf{73.01}  & \textbf{49.27}  & \textbf{35.46}  & \textbf{32.66}  & \textbf{64.83}  & \textbf{39.78}  & \textbf{36.41}  \\
    \hline
    FedAvg \cite{FedAVG} & 27.51  & 72.26  & 29.98  & 19.79  & 19.07  & 32.95  & 21.05  & 15.61  & 26.24  & 59.31  & 33.52  & 30.24  \bigstrut[t]\\
    +AGW$^\dagger$ \cite{AGW}  & 21.65  & 63.13  & 23.25  & 13.45  & 14.17  & 23.84  & 15.89  & 12.10  & 24.31  & 54.51  & 30.66  & 27.19 \\  
    +DNS$^\dagger$ \cite{DNS}  & 39.60  & 81.96  & 40.09  & 27.64  & 48.48  & 69.76  & 45.30  & 31.51  & 30.79  & 62.29  & 37.81  & 34.66  \\
    \rowcolor[rgb]{0.9, 0.9, 0.98}
    +DPPT (Ours) & \textbf{51.27}  & \textbf{88.55}  & \textbf{49.29}  & \textbf{34.47}  & \textbf{59.85}  & \textbf{77.58}  & \textbf{55.58}  & \textbf{41.70}  & \textbf{34.69}  & \textbf{67.20}  & \textbf{41.91}  & \textbf{38.48}  \\
    \hline
    \end{tabular}}
  \label{tab:1}%
\end{table*}

\begin{table*}[t]
  \centering
  \caption{Evaluation of our DPPT under EI and ES protocols. The upper part of the table is under ES, while the lower part is under EI. The underlined and bold indicate the best results for both protocols, respectively. B is the baseline using ERM with $\mathcal{L}_{id}$ and $\mathcal{L}_{cir}$ under ES, B$^\dagger$ denote the baseline using FedAvg supervised by $\mathcal{L}_{id}$ and $\mathcal{L}_{cir}$ under EI. We use R, L, and S to denote \textbf{R}egDB, \textbf{L}LCM, and \textbf{S}YSU-MM01 datasets, respectively. The left of $\rightarrow$ indicates seen entities and the right is the unseen entity.}
  \vspace{-1.0em}
  \resizebox{\textwidth}{!}{
    \begin{tabular}{c|cc|cccc|cccc|cccc}
    \hline
    \multirow{2}[4]{*}{Methods} & \multirow{2}[4]{*}{Param.} & \multirow{2}[4]{*}{FLOPs} & \multicolumn{4}{c|}{R \cite{RegDB} \textbf{+} L \cite{DEEN} $\mathbf{\rightarrow}$ S \cite{SYSU}} & \multicolumn{4}{c|}{L \cite{DEEN} \textbf{+} S \cite{SYSU} $\mathbf{\rightarrow}$ R \cite{RegDB}}    & \multicolumn{4}{c}{R \cite{RegDB} \textbf{+} S \cite{SYSU} $\mathbf{\rightarrow}$ L \cite{DEEN}} \bigstrut\\
\cline{4-15}          &       &       & r=1 $\uparrow$   & r=10 $\uparrow$  & mAP $\uparrow$   & mINP $\uparrow$  & r=1 $\uparrow$   & r=10 $\uparrow$  & mAP $\uparrow$   & mINP $\uparrow$  & r=1 $\uparrow$   & r=10 $\uparrow$  & mAP $\uparrow$   & mINP $\uparrow$ \bigstrut\\
    \hline
    B & 23.50  & 10.34  & 8.63  & 36.26  & 9.57  & 4.01  & 17.08  & 34.32  & 17.69  & 11.15  & 8.74  & 26.83  & 12.23  & 9.78  \bigstrut[t]\\
    LBA \cite{Lba}   & 23.55  & 10.36  & 8.09  & 34.63  & 9.55  & 4.29  & 12.01  & 28.65  & 11.69  & 6.01  & 8.38  & 26.80  & 12.20  & 9.90  \\
    AGW \cite{AGW}   & 23.55  & 10.36  & 9.59  & 38.28  & 10.42  & 4.45  & 13.79  & 29.76  & 14.05  & 8.25  & 9.08  & 26.77  & 12.37  & 9.80  \\
    DEEN \cite{DEEN}  & 41.23  & 27.70  & 9.48  & 38.39  & 10.03  & 3.72  & $\underline{19.42}$  & 37.59  & $\underline{19.44}$  & $\underline{12.33}$  & 10.50  & 29.41  & 14.09  & 11.48  \\
    CAJ \cite{CAJ}  & 23.55  & 10.36  & 10.90  & 40.57  & 11.18  & 4.39  & 16.55  & 37.23  & 17.40  & 10.74  & $\underline{11.35}$  & $\underline{31.17}$ & $\underline{15.03}$  & $\underline{12.10}$  \\
    DNS \cite{DNS}  & 25.45  & 10.36  & $\underline{11.75}$  & $\underline{42.36}$  & $\underline{11.77}$  & $\underline{4.56}$  & 18.87  & $\underline{37.79}$  & 18.36  & 10.56  & 10.14  & 28.47  & 13.55  & 10.94  \\
    \hline
    B$^\dagger$ & 23.50  & 5.17  & 9.72  & 38.21  & 10.74  & 4.75  & 15.37  & 29.95  & 16.77  & 10.90  & 8.97  & 26.81  & 12.85  & 10.62  \bigstrut[t]\\
    B$^\dagger$+CA & 23.50  & 5.17  & 10.10  & 39.34  & 10.73  & 4.19  & 17.61 & 33.58  & 17.73  & 11.31  & 14.40  & 35.87  & 18.89  & 16.30  \\
    DPPT (Ours) & 23.50  & 5.17  & \textbf{11.27}  & \textbf{41.20}  & \textbf{11.86}  & \textbf{5.34}  & \textbf{21.54}  & \textbf{40.37}  & \textbf{20.72}  & \textbf{12.78}  & \textbf{14.63}  & \textbf{37.03}  & \textbf{19.15}  & \textbf{16.44} \\
    \hline
    \end{tabular}}
  \label{tab:2}
  \vspace{-1.0em}
\end{table*}

\subsection{Experimental Settings}
\textbf{Datasets.} We conduct experiments on three widely-used VI-ReID datasets to evaluate our proposed method, i.e., SYSU-MM01 \cite{SYSU}, RegDB \cite{RegDB}, and LLCM \cite{DEEN}. 

\textbf{SYSU-MM01} contains 44,754 images of 491 identities captured by 4 visible and 2 infrared cameras, including 29,033 visible images and 15,715 infrared images. \textbf{RegDB} contains 412 identities collected by 1 visible and 1 infrared camera. Each identity has 10 visible and 10 infrared images. \textbf{LLCM} contains 46,767 images of 1,064 identities captured by 9 visible and infrared cameras, with the training and testing sets split at an approximate ratio of 2:1.

\noindent\textbf{Evaluation Protocols.} Following existing VI-ReID settings \cite{DEEN}, we adopt the rank-\textit{k} matching accuracy, mean Average Precision (mAP), and mean Inverse Negative Precision (mINP) \cite{AGW} as evaluation metrics. The protocol in terms of CI is defined in Sec.~\ref{sec:3.2}, and the protocols of EI and ES are defined in Sec.~\ref{sec:3.3}.
Following \cite{AGW,CAJ,DEEN,DNS,SYSU}, all datasets are used in infrared-to-visible mode for clarity, where infrared images serve as the query set, and visible images as the gallery. \textit{All the reported results are repeated ten times for a fair comparison.}

\subsection{Implementation Details} We adopt the ResNet-50 \cite{resnet} pretrained on the ImageNet-1k \cite{Imagenet} as our backbone. During the training state, all the input images are resized to $288\times 144$. The batch-size is set to 64, where randomly sampling 8 identities with 8 images. For the optimization, the Stochastic Gradient Descent (SGD) optimizer is adopted, with a weight decay of $5e^{-4}$ and momentum of 0.9. The initial learning rate is set to 0.2, and the OneCycleLR scheduler \cite{OneCycleLearning} is adopted. The epochs for CI and EC experiments are set to 50 and 30, respectively. The FedAvg \cite{FedAVG} is employed as our default. The TOP $K=4$ and $\lambda=1$ are decided by ablations on settings.

For CI, each client model is trained using data from a single camera within the training set of a given dataset. The global model is then evaluated on the testing set of the same dataset. For EI, we have two client models. Two datasets (entities) are used to train the two client models, respectively, and another new entity is used to test the global model. For ES, which is similar to EI, two entities are merged to train a model, and the model is tested on another new entity, which is not learned during training.

\begin{figure*}[t]
    \centering
    \includegraphics[width=0.95\textwidth]{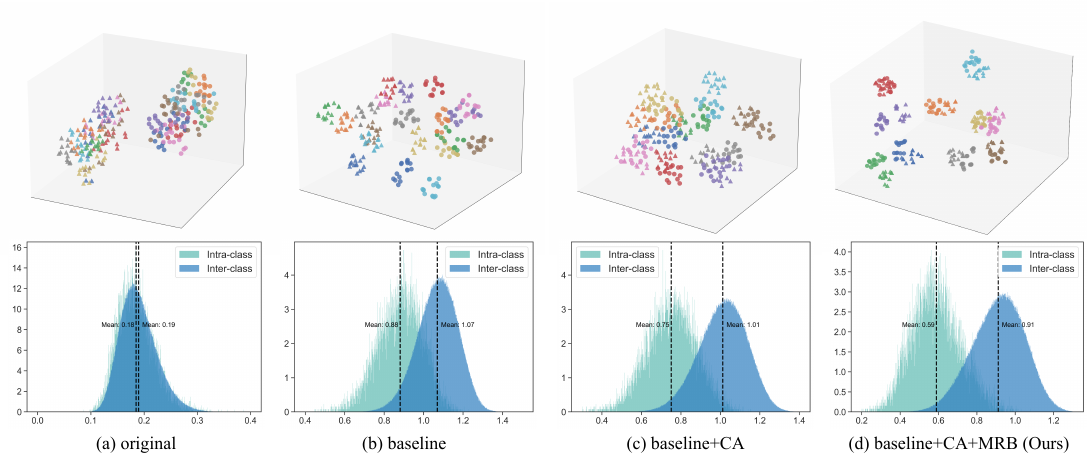}
    \vspace{-1.0em}
    \caption{t-SNE and intra-inter distances visualization of each component. The circles and triangles denote the visible and infrared modality, respectively. Different colors denote different identities.}
    \label{fig:3}
    \vspace{-1.0em}
\end{figure*}

\subsection{Qualitative Analysis}
\label{sec:4.3}

\textbf{Evaluation on the CI Protocol.} The results are reported on Tab.~\ref{tab:1}. It is noted that existing VI-ReID methods require paired visible and infrared images as input, making them unsuitable for addressing the modality incomplete challenge under the CI protocol. Therefore, we report our results using four classic federated learning algorithms: FedProx \cite{FedProx}, Fednova \cite{Fednova}, Moon \cite{MOON}, and FedAvg \cite{FedAVG}, each adopting different aggregation strategies. For these methods, the setting of local training is the same, conducting ResNet-50 supervised by $\mathcal{L}_{id}$ and $\mathcal{L}_{cir}$. The comparison methods after + only modify local training stage, with all other steps remaining consistent. The compared method only changes the local training stage. It is obvious that our proposed DPPT achieves consistent improvements across the board. Additionally, we modify two VI-ReID methods, i.e., AGW \cite{AGW} and DNS \cite{DNS}, by converting its two-stream architecture into a single-stream structure, adopting our proposed sampling approach, and removing its modality information (denoted by AGW$^\dagger$ and DNS$^\dagger$). They remain a significant performance gap compared with our DPPT.

\textbf{Evaluation on ES and EI Protocols.} We first replicate several recent VI-ReID methods \cite{Lba,AGW,DEEN,CAJ,DNS} to evaluate their generalization ability under the ES protocol, shown in Tab.~\ref{tab:2}. While most methods slightly outperform our baseline B (ERM with $\mathcal{L}_{id}$ and $\mathcal{L}_{cir}$), the differences are minimal, and overall rank-1 accuracy remains low. This suggests that current VI-ReID methods still have limited capability in handling unseen environments.
We encourage researchers to explore the bottlenecks that limit the generalization ability of existing methods, bringing VI-ReID closer to real world and advancing the community.
Moreover, we report our method under the EI protocol. Remarkably, our DPPT achieves comparable or even superior performance under EI compared to methods evaluated under ES. Moreover, The baseline results under the EI protocol (B$^\dagger$) are similar to those under the ES protocol (B), indicating that decentralized training does not significantly impact the model's ability to generalize to unseen environments.

\begin{table}[t]
  \centering
  \caption{Ablation studies on SYSU-MM01 and LLCM datasets. The \colorbox{rgb,255:red,230; green,230; blue,250}{light purple} is our final choice setting.}
  \vspace{-1.0em}
  \resizebox{\columnwidth}{!}{
    \begin{tabular}{c|ccc|ccc}
    \hline
    \multirow{2}[4]{*}{Settings} & \multicolumn{3}{c|}{SYSU-MM01 \cite{SYSU}} & \multicolumn{3}{c}{LLCM \cite{DEEN}} \bigstrut\\
\cline{2-7}          & r=1 $\uparrow$   & mAP $\uparrow$   & mINP $\uparrow$  & r=1 $\uparrow$   & mAP $\uparrow$   & mINP $\uparrow$ \bigstrut\\
    \hline
    B     & 27.51  & 29.98  & 19.79  & 26.24 & 33.52 & 30.24 \bigstrut\\
    \hline
    B+CA  & 39.47  & 39.78  & 26.95  & 30.47 & 37.30  & 34.09 \bigstrut[t]\\
    B+gray & 35.25  & 36.16  & 24.03  & 29.33 & 36.29 & 33.06 \\
    \hline
    \rowcolor[rgb]{0.9, 0.9, 0.98}
    B+CA+$\mathcal{L}_{mrb}$(cos) & \textbf{51.27}  & \textbf{49.29}  & \textbf{34.47}  & \textbf{34.69} & \textbf{41.91} & \textbf{38.48} \bigstrut[t]\\
    B+CA+$\mathcal{L}_{mrb}$(edu) & 43.79  & 44.00  & 30.77  & 33.45 & 40.62 & 37.35 \\
    B+CA+$\mathcal{L}_{mrb}$(mixed) & 47.82  & 47.55  & 33.94  & 34.41 & 41.56 & 38.09 \\
    \hline
    \end{tabular}}
  \label{tab:3}
  \vspace{-1.0em}
\end{table}

\subsection{Ablation Study}

\textbf{Effectiveness of Each Component.} To evaluate the contribution of each component, we conduct a series of ablation experiments on SYSU-MM01 \cite{SYSU} and LLCM \cite{DEEN} datasets Under CI protocol. The results are shown in Tab.~\ref{tab:3}. Baseline denotes that we employ the FedAvg \cite{FedAVG} to train the ResNet-50 and supervised by $\mathcal{L}_{id}$ and $\mathcal{L}_{cir}$. Compared with the grayscale, CA brings a superior improvement. This is because CA preserves original information by randomly mapping one channel to the other two, whereas grayscale discards valuable color information crucial for identity discrimination. Subsequently, our proposed $\mathcal{L}_{mrb}$ further improves the performance. In addition, to explore the role of cosine similarity in Eq.~\ref{eq:6}, we substituted it with Euclidean distance (edu) and a combination of both (mixed), as detailed in the supplementary material. The ablation study under EI is shown in Tab.~\ref{tab:2}, showing consistent improvement by adding each component.

\textbf{Visualization.} To demonstrate the effectiveness of each component, we visualize the distribution of learned features by t-SNE \cite{TSNE} and intra-class and inter-class distances, as shown in Fig.~\ref{fig:3}. Compared with (a) which only uses ResNet-50 pretrained on the ImageNet-1k, the baseline improves the discriminative of identity within the modality but still exhibits a significant cross-modality gap. Subsequently, the integration of CA effectively bridges this huge discrepancy, pushing away the intra-class and inter-class distances. Despite encouraging progress, the feature distribution map shows that the similarity between different identities is relatively close, making it hard to distinguish. Finally, with the help of MRB, the modality gap is further reduced. The reason can be attributed to that MRB handles the \textit{domain shift} and \textit{identity missing} well, thus achieving extremely smaller intra-class and giant inter-class distances, facilitating easier distinction between identity samples.

\begin{figure}[t]
    \centering
    \includegraphics[width=0.48\textwidth]{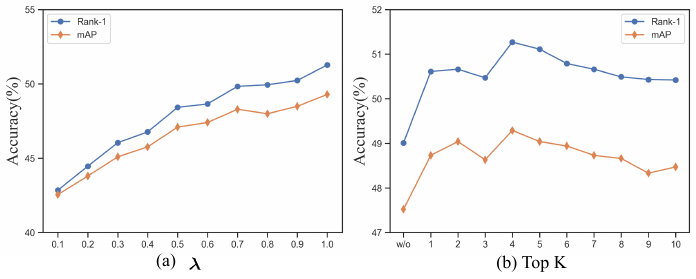}
    \vspace{-2.0em}
    \caption{The effect of hyperparameter $\lambda$ and \textit{K} on SYSU-MM01.}
    \label{fig:4}
    \vspace{-1.5em}
\end{figure}

\textbf{Hyper-Parameter Ablation.} The hyper-parameter ablation of $\lambda$ and Top $K$ is as shown in Fig.~\ref{fig:4}(a) and (b). The $\lambda$ is set to 1.0 and the $K$ is set to 4.

\section{Conclusion}
This paper presents L2RW, the first privacy-preserved benchmark for VI-ReID, aims at bringing VI-ReID closer to real-world applications. To simulate privacy constraints, we introduce three protocols: camera independence (CI), entity independence (EI) and entity sharing (ES). We also propose the DPPT method, which addresses challenges under both protocols without compromising data privacy. Extensive experiments validate the feasibility of privacy-preserved VI-ReID and demonstrate the effectiveness of our method which achieves significant improvement under privacy protection conditions. Our results further show that decentralized training does not significantly affect generalization. We hope this work inspires future research in privacy-preserved VI-ReID and advances the field.

\section{Acknowledgement}
This work was supported by the National Natural Science Foundation of China (Grant No.~U22B2062), the Research Council of Finland (former Academy of Finland) Academy Professor project EmotionAI (grants 336116, 345122, 359854), the University of Oulu \& Research Council of Finland Profi 7 (grant 352788), EU HORIZON-MSCA-SE-2022 project ACMod (grant 101130271), the Finnish Doctoral Program Network in Artificial Intelligence, AI-DOC (decision number VN/3137/2024-OKM-6), and the Startup Foundation for Introducing Talent of NUIST.


{
    \small
    \bibliographystyle{ieeenat_fullname}
    \bibliography{main}
}


\end{document}